\documentclass[runningheads]{llncs}
\usepackage{graphicx}

\usepackage{tikz}
\usepackage{comment}
\usepackage{amsmath,amssymb} %
\usepackage{color}
\usepackage{orcidlink}

\usepackage[accsupp]{axessibility}  %

\usepackage{hyperref}
\usepackage{tabularx}
\usepackage{booktabs}
\usepackage{caption} 
\usepackage[normalem]{ulem}
\usepackage{multirow}
\usepackage{adjustbox} %
\definecolor{green}{rgb}{0, 0.5, 0}
\definecolor{orange}{rgb}{0.6, 0.3, 0.1}
\definecolor{Red}{rgb}{1.0, 0.0, 0.0}
\definecolor{teal}{rgb}{0.0, 0.4, 0.4}
\definecolor{purple}{rgb}{0.65,0,0.65}
\definecolor{saffron}{rgb}{0.95,0.75,0.2}
\definecolor{turquoise}{rgb}{0.0,0.5,0.5}
\definecolor{brown}{rgb}{0.5, 0.16, 0.16}
\definecolor{brickred}{rgb}{.6, .2 .1}
\definecolor{coral}{rgb}{1,0.45,0.33}

\begin{document}
\pagestyle{headings}
\mainmatter
\def\ECCVSubNumber{3218}  %

\title{Capturing, Reconstructing, and Simulating:\\ the UrbanScene3D Dataset} %

\titlerunning{UrbanScene3D}
\author{Liqiang Lin\orcidlink{0000-0003-2594-6495} \and
    Yilin Liu\orcidlink{0000-0001-7336-1956} \and
    Yue Hu \and
    Xingguang Yan \and
    Ke Xie \and
    Hui Huang\thanks{Corresponding author}\orcidlink{0000-0003-3212-0544}%
}
\authorrunning{L. Lin et al.}
\institute{Shenzhen University}
\maketitle

\begin{abstract}
    We present \emph{UrbanScene3D}, a large-scale data platform for research of urban scene perception and reconstruction.
    UrbanScene3D contains over 128k high-resolution images covering 16 scenes including large-scale real urban regions and synthetic cities with 136 $km^2$ area in total.
    The dataset also contains high-precision LiDAR scans and hundreds of image sets with different observation patterns, which provide a comprehensive benchmark to design and evaluate aerial path planning and 3D reconstruction algorithms.
    In addition, the dataset, which is built on Unreal Engine and Airsim simulator together with the
    manually annotated unique instance label for each building in the dataset,
    enables the generation of all kinds of data, e.g., 2D depth maps, 2D/3D bounding boxes, and 3D point cloud/mesh segmentations, etc.
    The simulator with physical engine and lighting system not only produce variety of data but also enable users to simulate cars or drones in the proposed urban environment for future research.
    The dataset with aerial path planning and 3D reconstruction benchmark is available at: {\bf\href{https://vcc.tech/UrbanScene3D}{\color{magenta}{https://vcc.tech/UrbanScene3D}}}

\end{abstract}

\section{Introduction}

With the development of digital photography and 3D scanning technologies, we have witnessed the explosive growth of data in recent years. Rich data sources and interaction methods bring rapid research progress in computer vision, computer graphics and robotics. For indoor scenes, with the help of sufficient data and real-time interactions~\cite{gibson,s3dis,nyu_depth,partnet}, many fundamental problems, such as 2D/3D object detection and segmentation~\cite{maskrcnn,pointnetplus}, depth estimation~\cite{consistent_depth,depth_wei_2021}, 3D reconstruction~\cite{3dr2n2,AtlasNet,ChenZ19,bspnet,DISN} and autonomous navigation~\cite{zhu2017target,chen2019behavioral,gupta2017cognitive}, have been better solved in a data-driven manner. However, things are different for outdoor study. The lack of effective devices and the extensive scales dramatically increase the difficulty of outdoor data capturing~\cite{holicity}. Also, due to the varied weather and light conditions, the outdoor scenes change fast and thus pose additional challenges to structure the data and design robust acquisition algorithms.

\begin{figure}[!t]
    \centering
    \includegraphics[width=\linewidth]{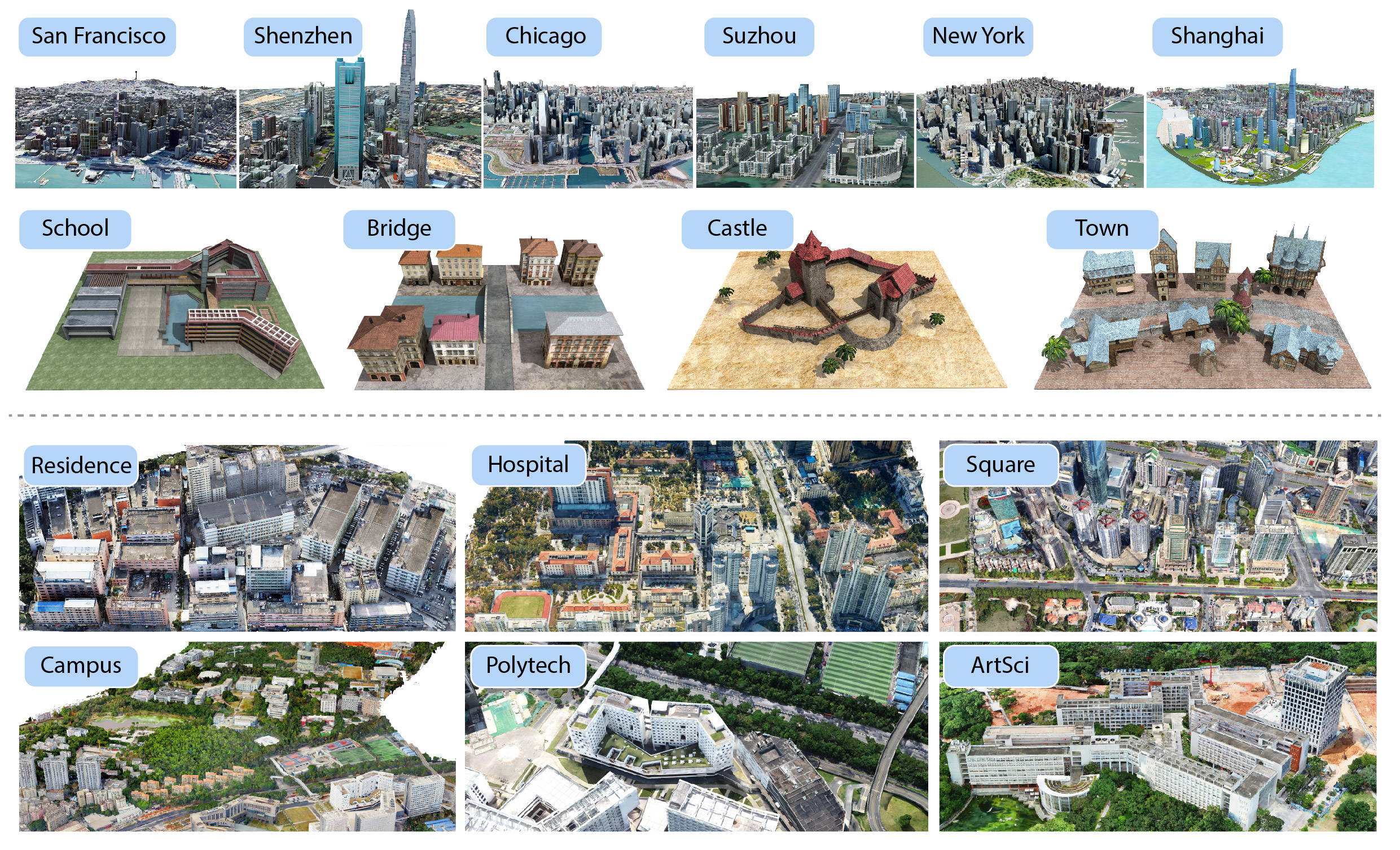}
    \caption{A glance of synthetic (top) and real (bottom) scenes in UrbanScene3D.
    }
    \label{fig:gallery}
\end{figure}

The current outdoor datasets are usually built by onboard equipment~\cite{kitti,cityscapes,apollo}, e.g., RGB cameras and/or LiDAR. 
Nonetheless, it is challenging to use these sensors to thoroughly capture the whole environment due to the limited field of views and routing choices. 
Other urban datasets~\cite{holicity,smith2018aerial} constructed by 3D modelers are usually clean and complete, but the models inside generally lack geometric and textural details. The gap between the proposed datasets and real-world scenarios remains large.

To tackle these problems, we present a large-scale urban scene dataset, \emph{UrbanScene3D}, which consists of both man-made and real-world reconstruction scenes in different scales, together with a convenient simulator built on Unreal Engine and AirSim. The man-made scene models have compact structures, which are carefully constructed/designed by professional artists; 
see the top of Fig.~\ref{fig:gallery}. Probably more important, UrbanScene3D also offers dense, detailed scene models reconstructed by aerial images through multi-view stereo (MVS) techniques; check out the bottom of Fig.~\ref{fig:gallery}, where the scene models are with realistic textures and meticulous geometric structures.

In particular, to investigate how to better acquire and reconstruct outdoor scenes, we select a set of scene representatives and capture them with a drone flying along a set of aerial paths. These flights are calculated by different planning algorithms for 3D urban scene reconstruction. Thus, for each representative environment, we are able to provide a variety of reconstructed meshes with the corresponding scene observations (aerial acquisition paths and the captured image sets). Besides, with the help of a high-precision laser scanner applied in the real world and the synthetic ground-truth models, we have constructed a benchmark that provides the point-level accuracy and completeness analysis of each reconstructed mesh. This enables a robust evaluation of both path planning strategies and MVS algorithms. In addition, the physical engine of AirSim enables users to simulate the robots (cars/drones) and test a variety of autonomous tasks in the proposed environments. The involvement of both synthetic and real scenes effectively extends the generalization ability of resulting algorithms.

In summary, our main contributions include: i) a large-scale urban scene dataset (Sec.~\ref{sec:dataset}) that facilitates research in various areas; ii) a comprehensive benchmark (Sec.~\ref{sec:evaluation}) for investigating the impact of different factors in aerial path planning (Sec.~\ref{sec:planner}) for 3D urban reconstruction; iii) an easy-to-use simulation platform (Sec.~\ref{sec:app}) for autonomous driving, robotics, and embodied AI research.

\section{Related Work}
\label{sec:related}

\paragraph{Outdoor datasets.}
The fast development of autonomous driving involves enormous outdoor datasets~\cite{apollo,kitti,sun2020scalability,behley2019semantickitti,cityscapes}. They often offer stereo sequences, 3D LiDAR point clouds, camera calibration, and 3D object tracklet labels for outdoor scenes, thus promoting many applications and research.
Although these ground-based sensors can capture small-scale scenes~\cite{Knapitsch2017}, they usually have very restricted views and routing choices, which cause significant challenges to cover large-scale urban areas.
Meanwhile, the unmanned aerial vehicles (UAVs) or drones, have much better visibility and more freedom to maneuver, making them suitable for a complete coverage of wide regions.
However, most existing UAV based datasets are not designed for capturing the entire scene. Instead, they only provide partial observations for perception tasks, such as semantic segmentation~\cite{lyu2020uavid}, object detection~\cite{du2018unmanned,zhu2019visdrone,mandal2020mor}, action recognition~\cite{chaquet2013survey}, gesture recognition~\cite{pisharady2015recent,perera2018uav}, or tracking~\cite{bergmann2019tracking,mueller2016benchmark}.
On the contrary, we carefully plan the drone path to obtain a complete capture, from which the entire 3D scene can be reconstructed with MVS methods.

\paragraph{Synthetic CAD datasets.}
Different from real-world datasets, building synthetic datasets with CAD models~\cite{liu2020novel,brunel2021flybo} can offer a complete structured environment at a much lower cost, but lack geometric and textural details.
To bridge the gap, HoliCity~\cite{holicity} aligns the real-world panoramas with the CAD models to provide real texture.
However, the discrepancy is still significant since the geometry is too coarse and the panoramas only covers a portion of the entire scene.
Instead, our UrbanScene3D contains both real-world and synthetic CAD scenes, facilitating research for both high-quality urban reconstruction and holistic scene understanding. 
Similar to Mueller et al.~\cite{mueller2016benchmark}, we also provide a simulator that stimulates the research of online real-time capturing and understanding of 3D urban scenes.

\paragraph{Aerial path planning for urban scene capture.}
To capture urban scenes with drones, aerial path planning plays a vital role; see a recent survey~\cite{zhou2020survey}.
Manual control and Zig-Zag patterns are inefficient, difficult to achieve decent coverage, and challenging to fulfill practical factors, such as safety restriction and battery capacity.
To deal with these problems, existing methods~\cite{schmid2012view,roberts2017submodular,hepp2018plan3d,smith2018aerial,koch2019automatic,zhou2020offsite,zhang2021continuous} optimize the drone path according to certain goals and constraints with a coarse proxy model or a top view image as input.
Specifically, Smith et al.~\cite{smith2018aerial} designs an optimization objective for better multi-view stereo results, ensuring the completeness and accuracy of the reconstruction.
Further, Zhang el al.~\cite{zhang2021continuous} propose a continuous path planner to adequately shorten the path length and reduce sharp turns.
To be able to do offsite planning, Zhou et al.~\cite{zhou2020offsite} utilize a satellite image to estimate a 2.5D proxy for view selection. 
For online planning, the researchers learn to construct 2.5D height maps~\cite{liu2021vgf} or estimate 3D bounding boxes of buildings~\cite{liu2021aerial} on-the-fly for drone navigation and exploration.
Due to the lack of valid data, these methods usually rely on a handcraft heuristic function to optimize the capturing views or the flight trajectories. However, the quality of the final reconstruction is constrained by the high-order relation among observations, which is difficult to model by heuristics and optimization designs.
Our data and benchmark will efficaciously facilitate future research on this topic.
\section{The UrbanScene3D Dataset}
\label{sec:dataset}

\begin{table}[!t]
    \footnotesize
    \caption{
        Statistics of UrbanScene3D with 10 synthetic and 6 real scenes.
        Area: the covered area of the scene model;
        Size: the size of the scene model;
        Texture: the number of texture images contained in the scene model;
        Texture: the size of texture;
        Object: the number of objects in the scene model.}
    \label{tab:datasets}
    \begin{tabularx}{\textwidth}{cc*{5}{>{\centering\arraybackslash}X}}
        \toprule
        Scene         & Area$(m^2)$      & Size$(Mb)$ & Texture(\#) & Texture$(Mb)$ & Object(\#) \\
        \midrule
        New York      & $7.4\times 10^6$ & 86         & 762         & 122           & 744        \\
        Chicago       & $2.4\times 10^7$ & 146        & 2277        & 227           & 1629       \\
        San Francisco & $5.5\times 10^7$ & 225        & 2865        & 322           & 2801       \\
        Shenzhen      & $3.0\times 10^6$ & 50         & 199         & 73            & 1126       \\
        Suzhou        & $7.0\times 10^6$ & 191        & 395         & 24            & 168        \\
        Shanghai      & $3.7\times 10^7$ & 308        & 2285        & 220           & 6850       \\
        School        & $1.7\times 10^4$ & 25         & 47          & 488           & 3          \\
        Bridge        & $1.3\times 10^4$ & 28         & 237         & 44            & 8          \\
        Castle        & $7.0\times 10^3$ & 9          & 47          & 184           & 6          \\
        Town          & $4.4\times 10^3$ & 112        & 73          & 348           & 17         \\
        \midrule
        Campus        & $1.3\times 10^6$ & 1859       & 122         & 3676          & 178        \\
        Residence     & $1.0\times 10^5$ & 356        & 52          & 1760          & 34         \\
        Square        & $1.1\times 10^6$ & 3665       & 799         & 980           & 156        \\
        Hospital      & $5.0\times 10^5$ & 6266       & 94          & 744           & 114        \\
        Polytech      & $1.5\times 10^4$ & 3523       & 50          & 221           & 3          \\
        ArtSci        & $1.6\times 10^4$ & 25395      & 118         & 556           & 3          \\
        \bottomrule
    \end{tabularx}
\end{table}

\begin{table}[!t]
    \footnotesize
    \caption{
        More path planning statistics of 4 synthetic and 2 real representative scenes selected from UrbanScene3D.
        Tri: the number of triangles of the scene model;
        Proxy: the number of different levels of proxies provided;
        Overlap: the overlap rate used to sample the proxies;
        Planner: the number of different planners used to generate aerial paths;
        Path: the total number of generated flight paths of the scene;
        Image: the total number of captured images of the scene.
    }
    \label{tab:planner}
    \centering
    \begin{tabularx}{0.95\textwidth}{c*{7}{>{\centering\arraybackslash}X}}
        \toprule
        Scene    & Tri(\#)   & Proxy(\#) & Overlap(\%) & Planner(\#) & Path(\#) & Image(\#) \\
        \midrule
        School   & 250,282   & 4         & 70 \& 90    & 4           & 25       & 14,897    \\
        Bridge   & 394,022   & 4         & 70 \& 90    & 4           & 25       & 13,228    \\
        Castle   & 109,513   & 4         & 70 \& 90    & 4           & 25       & 7,414     \\
        Town     & 1,197,751 & 4         & 70 \& 90    & 4           & 25       & 8,948     \\
        \midrule
        Polytech & 2,613,608 & 3         & 90          & 4           & 13       & 19,635    \\
        ArtSci   & 4,524,028 & 3         & 90          & 4           & 13       & 12,261    \\
        \bottomrule
    \end{tabularx}
\end{table}

The goal of UrbanScene3D is to provide a general data platform for 3D vision, graphics and robotics research in urban scene environments with different scales.
UrbanScene3D provides 10 synthetic and 6 real-world scenes with CAD and reconstructed mesh models and the corresponding aerial images; see Fig.~\ref{fig:gallery} and Table~\ref{tab:datasets} for more details.
The synthetic CAD scenes consist of various compact primitive structures including buildings, bridges, streets, and vegetation, all of which are built by professional artists.
For real scenes, we use a drone to capture images for MVS.
We program the drone to follow an aerial path generated with oblique photography, a commonly used industrial solution conducted by \emph{DJI-Terra}\footnote{https://www.dji.com/dji-terra} for 3D city acquisition.
Based on these images, we reconstruct the scenes with \emph{ContextCapture}\footnote{https://www.bentley.com/en/products/brands/contextcapture}, a commercial MVS solution.
Specially, the selected representative scenes for our benchmark include additional huge volume of capture data using various path planners under different settings; see Table~\ref{tab:planner}.

For oblique photography planner, we use a professional \textit{DJI M300RTK}\footnote{https://www.dji.com/matrice-300} drone loaded with five HD \textit{PSDK 102S} aerial cameras. %
For the other planners, we use a single-camera \textit{DJI PHANTOM 4 RTK}\footnote{https://www.dji.com/phantom-4-rtk} drone.%

\begin{table}[!t]
	\footnotesize
	\caption{
		Comparing UrbanScene3D with existing 3D outdoor datasets. Area stands for the maximum area across all scenes in the dataset.
		Path stands for the number of flights/rides for capturing. Note that for BlendedMVS, we only show the estimated statistics of the urban scenes.}

	\label{tab:comparison}
	\begin{adjustbox}{width=\columnwidth,center}
		\setlength{\tabcolsep}{4pt}
		\begin{tabular}{cccccccccc}
			\toprule
			Datasets                                & Scene   & Type     & Area($km^2$) & Path(\#) & Image(\#) & Diversity        & Simulator  & Semantics & \\[0.4ex] \midrule
			KITTI~\cite{kitti}                      & driving & LiDAR    & /            & 1        & 93k       & 5 scans          & /          & /         & \\[0.4ex]
			Cityscape~\cite{cityscapes}             & driving & stereo   & /            & /        & 25k       & 50 cities        & /          & 2D        & \\[0.4ex]
			HoliCity~\cite{holicity}                & urban   & CAD      & 20           & /        & 6.3k      & 1 city           & /          & 3D        & \\[0.4ex]
			BlendedMVS~\cite{yao2020blendedmvs}     & mixed   & MVS      & 0.02         & 29       & 17k       & 29 scenes        & /          & /         & \\[0.4ex]
			SYNTHIA~\cite{ros2016synthia}           & driving & CAD      & /            & /        & 50k       & 1 city           & car        & 3D        & \\[0.4ex]
			Mueller2016~\cite{mueller2016benchmark} & mixed   & CAD      & /            & /        & /         & /                & drone      & 3D        & \\[0.4ex]
			Smith2018~\cite{smith2018aerial}        & urban   & CAD      & 0.03         & /        & /         & 5 scenes         & /          & /         & \\[0.4ex]
			\midrule
			\textbf{UrbanScene3D}                   & urban   & CAD\&MVS & 55           & 130      & 128k      & 16 cities/scenes & car\&drone & 3D        & \\ \bottomrule
		\end{tabular}
	\end{adjustbox}
\end{table}

Table~\ref{tab:comparison} summarizes the difference between UrbanScene3D and the existing outdoor datasets.
We further highlight the features of UrbanScene3D as follows.

\paragraph{Up in the air.}
Most outdoor datasets are ground based and the capture usually do not cover the entire scene.
The capture is incomplete even for aerial dataset like UAVDet~\cite{du2018unmanned}, since its goal is object detection.
Similar to BlendedMVS~\cite{yao2020blendedmvs}, we provide aerial captures that are suitable for MVS.
Specifically for the representative scenes, we adopt multiple optimized aerial paths for the capture, which lead to greater quality urban scenes.
Our approach not only provides a high-quality dataset for local perception research but also enables the research of global understanding for real 3D urban scenes.

\paragraph{Extensive scale.}
Many existing datasets do not offer the complete capture of large-scale scenes; see Table~\ref{tab:comparison}.
While Holicity offers a $20 km^2$ scene of London, our dataset includes three large-scale city scenes that cover above $24 km^2$ area.
Meanwhile, we offer multiple real-world complete scenes, which include two extensive scenes that cover above $1 km^2$ area, whose scale is unseen in previous datasets like BlendedMVS~\cite{yao2020blendedmvs}.
More diverse urban scenes and the corresponding aerial images with shot poses are also available.
Additionally, our drone flight path length can be up to $17km$, which benefits the research for SLAM or SfM.

\paragraph{Path planning research.}
For current path planning methods, factors such as the proxy accuracy and overlap rate play an vital role.
Our path planning benchmark specifically include different settings for these factors.
We show their influence on the final reconstruction quality and acquisition efficiency; see Sec.~\ref{sec:evaluation}.
In contrast, existing benchmark~\cite{smith2018aerial} for path planning does not consider these factors and only includes synthetic scenes.
Besides, our benchmark includes hundreds of flight paths, which will boost the future research for path planning techniques.

\paragraph{Multiple captures.}
The different weather and lighting conditions in outdoor environment pose huge challenges for perception and reconstruction in general.
Our benchmark offers multiple drone flights for each scene over different times, this greatly increases the variety of data, which can benefit learning based methods to in turn solve this problem.
For example, the abundance (10k+) of the real scene images in the benchmark under different lighting conditions allows the future research for NeRFs~\cite{mildenhall2020nerf} that can decouple geometry, material and light.

\paragraph{Simulation environment.}
The sim-to-real gap is a critical problem for the robotics and embodied AI research.
Our simulator can directly import our real-world scenes and simulate the drones inside them, hence this gap would be greatly shortened.
In contrast, most existing simulator for autonomous driving or UAVs operate only for virtual scenes. Further, our simulator can show the coverage of the scene in real-time, which is useful for research on UAV exploration.

\section{Scene Acquisition with Aerial Path Planning}
\label{sec:planner}

UrbanScene3D offers a wide variety of potential applications, from instance segmentation, multi-view stereo, to depth estimation and novel view synthesis. See Sec.~\ref{sec:app} for further details. 
In this section, we extend the UrbanScene3D dataset to a benchmark to evaluate the quality and efficiency of different path planning algorithms, as well as the impact of input proxies in Sec.~\ref{sec:evaluation}.

UrbanScene3D contains all the data needed for testing path planning algorithms and analyzing the reconstructed results, including proxies, ground-truth models/point clouds, paths, images, and reconstructed results; see Table~\ref{tab:planner}.

The majority of path planners usually rely on a pre-computed coarse model (also called \textit{proxy}~\cite{smith2018aerial}) of the environment. The proxy can be obtained by various methods, including a quick reconstruction after a simplified oblique photography pass~\cite{smith2018aerial,roberts2017submodular}, satellite image~\cite{zhou2020offsite}, map providers or even through a real-time reconstruction~\cite{liu2021aerial}. The quality of the proxy, including the accuracy of the topology and the face normal, can affect the final quality of the reconstruction.

In Sec.~\ref{sec:benchmark:planner}, we briefly introduce the 4 path planners we used to generate different trajectories. To evaluate the influence of the proxy in the path planning, we then introduce different proxies we used in the dataset in Sec.~\ref{sec:benchmark:proxies}.

\subsection{Aerial Path Planning Methods}
\label{sec:benchmark:planner}

\begin{figure}[!t]
    \centering
    \includegraphics[width=\linewidth]{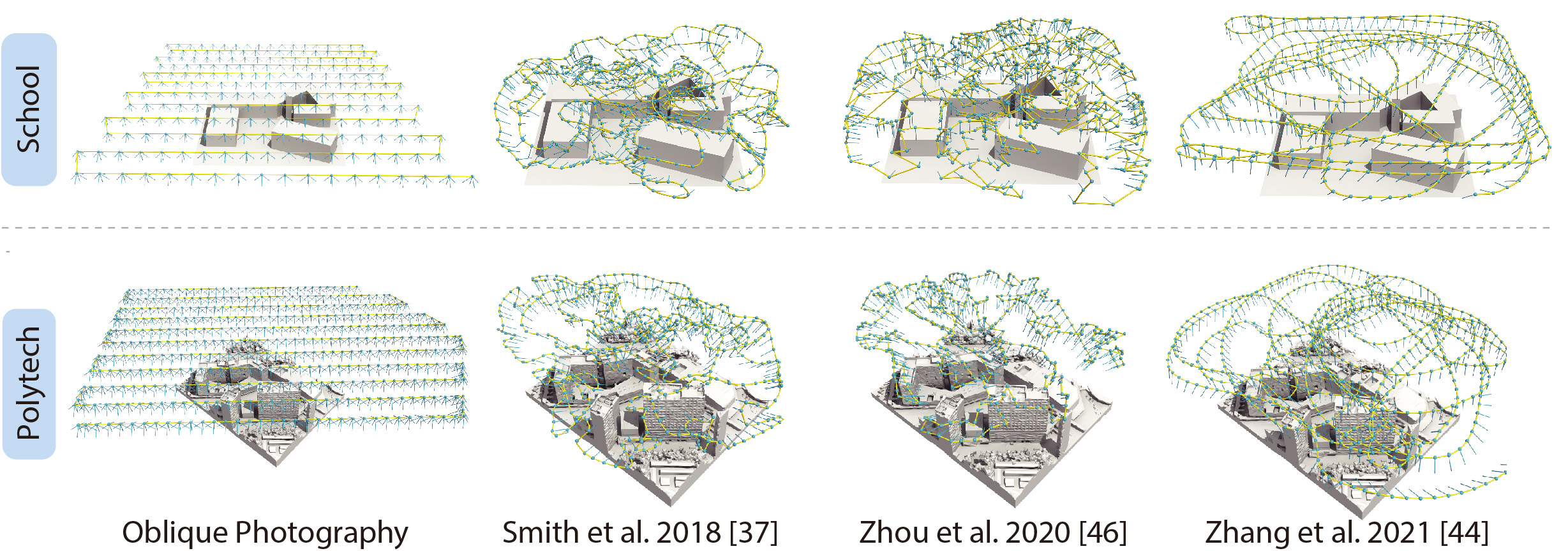}
    \caption{
        The comparison of the paths generated with different aerail planning methods on the synthetic scene School and the real scene Polytech.
    }
    \label{fig:paths}
\end{figure}

In order to offer more paths and images, we use four different path planners, including oblique photography and the methods proposed by Smith et al.~\cite{smith2018aerial}, Zhou et al.~\cite{zhou2020offsite}, and Zhang et al.~\cite{zhang2021continuous}, to generate the paths on the different proxies mentioned in Sec.~\ref{sec:benchmark:proxies} with different overlap rates for different scenes, resulting in 100 different paths for synthetic scenes and 26 paths for real scenes. Fig.~\ref{fig:paths} shows an illustration of the paths generated with the four different planners on the synthetic scene School and the real scene Polytech.

\paragraph{Oblique photography.}
Given the image overlap and ground sample distance (GSD), Oblique photography generate an S-shaped trajectory at a fixed height (calculated by the GSD) and compute the required capture location.
The S-shaped trajectory is usually computed by \textit{Complete Coverage Path Planning} (CCPP) algorithm, which could guarantee the complete coverage of an area even with an irregular shape~\cite{ZHANG2020198}. Also, the number of turns is minimized, which could significantly increase the capturing efficiency.

\paragraph{Planner proposed by Smith et al.~\cite{smith2018aerial}.}
Unlike Oblique photography, Smith et al.~\cite{smith2018aerial} directly optimize viewpoints according to a heuristic function, \emph{reconstructability},
which consider both the potential error of triangulation and feature matching in the \textit{Multi-view Stereo} (MVS) process. In each iteration, Smith et al.~\cite{smith2018aerial} first compute the so-called \emph{reconstructability}
of each point respect to the current viewpoint set. Then they try to maximize this measurement by adjusting the position and orientation of each viewpoint. 

\paragraph{Planner proposed by Zhou et al.~\cite{zhou2020offsite}.}
Similar to Smith et al.~\cite{smith2018aerial}, Zhou et al.~\cite{zhou2020offsite} also consider the reconstructability
of each point during the planning. However, they only use this measurement to reduce useless viewpoints. They first generate a large viewpoint set, which suppose to be \textit{complete} but highly \textit{redundant}. In each iteration, they define the \textit{view redundancy} on the computed reconstructability and delete the most redundant viewpoint accordingly. 

\paragraph{Planner proposed by Zhang et al.~\cite{zhang2021continuous}.}
Compared with Oblique photography, Smith et al.~\cite{smith2018aerial} and Zhou et al.~\cite{zhou2020offsite} generate trajectories with higher capturing quality. However, the heuristic function which is defined on the viewpoints brings many sharp turns in the final trajectories. The drastic speed change significantly decreases the capturing efficiency. Thus, Zhang et al.~\cite{zhang2021continuous} involves path smoothness in the heuristic function and utilize \textit{Rapidly-exploring random} (RRT) tree to search for an efficient and high-quality trajectory.

\subsection{Geometric Proxies}
\label{sec:benchmark:proxies}

\begin{figure}[!t]
    \centering
    \includegraphics[width=\linewidth]{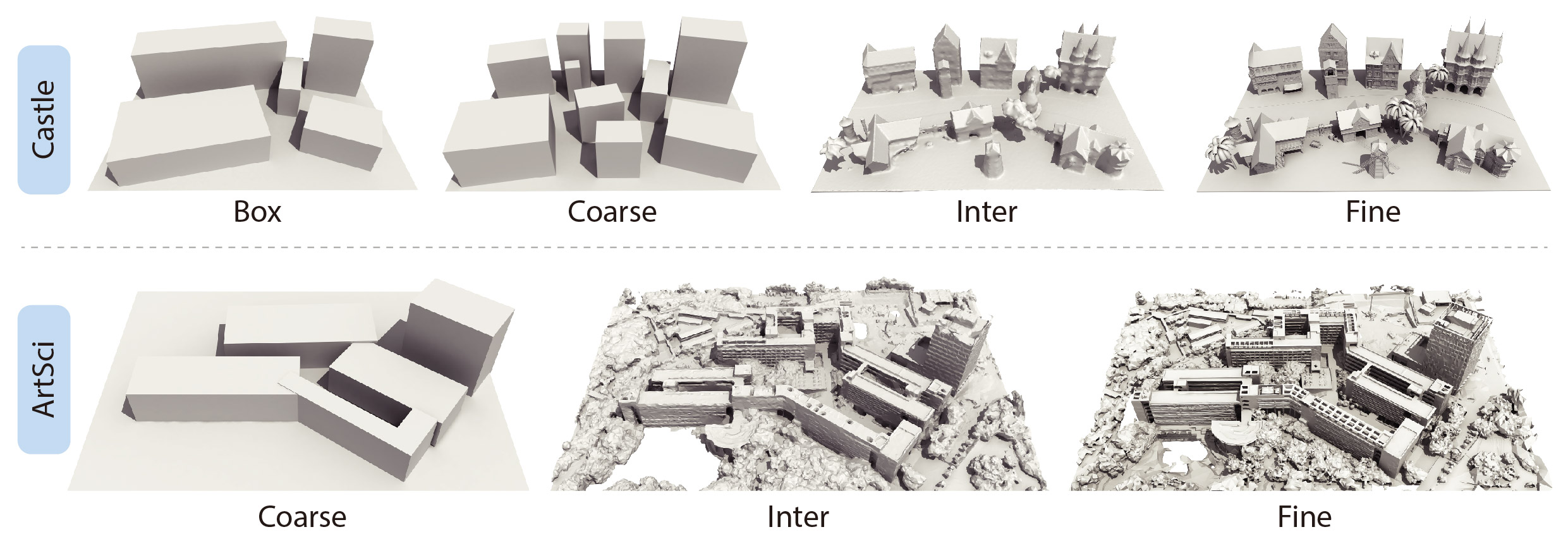}
    \caption{
        The proxies of scene School in different levels of details.
        Box: the roughest proxy;
        Coarse: the coarse level of proxy;
        Inter: the intermediate level of proxy;
        Fine: the finest level of proxy.
    }
    \label{fig:proxies}
\end{figure}

The proxies are essential for aerial path planning methods. Detailed proxy usually leads to much better reconstruction results.
Previous works either use a rough scene proxy~\cite{zhang2021continuous,smith2018aerial} or a 2.5D model extracted from satellite images~\cite{zhou2020offsite} to plan the path. UrbanScene3D provides proxies in different levels of details, which could be used to analyze the relations between proxies, paths, and the corresponding qualities of the reconstructed models; see Fig.~\ref{fig:proxies}.

The box proxy (\textit{box}) is the roughest level of proxy with incorrect topology.
It is built by replacing the building set in the scene with its bounding box. For the real scene, the box proxy is deprecated due to the safety issue.
Similar to \textit{box} proxy, \textit{coarse} proxy is also built by finding the bounding box of each building in the scene. However, the \textit{coarse} proxy has 
more accurate topology, which might lead more accurate path planning result.
The intermediate proxy (\textit{inter}) is reconstructed by downsampled images which are captured by oblique photography.
The ground-truth meshes (\textit{fine}) for the synthetic scenes can also be used as proxies, which is supposed to provide the largest geometric information during the path planning process.
The \textit{fine} proxy for the real scenes are reconstructed with non subsampled images captured by oblique photography.
Since there are no ground-truth meshes for real scenes, we use reconstructed models by high-density images as their corresponding fine proxies.

\section{Scene Reconstruction Benchmarks}
\label{sec:evaluation}

In this section, we evaluate the paths generated with different path planning methods for both energy cost, aerotriangulation accuracy, and reconstruction quality, providing the point-level accuracy and completeness analysis of the reconstructed mesh.
The statistics of energy consumption of UAV capturing is given in Sec.~\ref{sec:benchmark:cost}.
In Sec.~\ref{sec:benchmark:reg}, we analyze the aerotriangulation results of different planners.
And finally, we evaluate the reconstruction results in Sec.~\ref{sec:benchmark:recon}. The Sec.~\ref{sec:benchmark:conclusion} give a overall comparison of all the four planners.

Other information, e.g., the evaluation of different overlaps, the cost of model reconstruction, and the reconstruction evaluation on the other scenes are included in the supplementary material.

\subsection{High-precision LiDAR Scan}

\begin{figure}[!t]
    \centering
    \includegraphics[width=\linewidth]{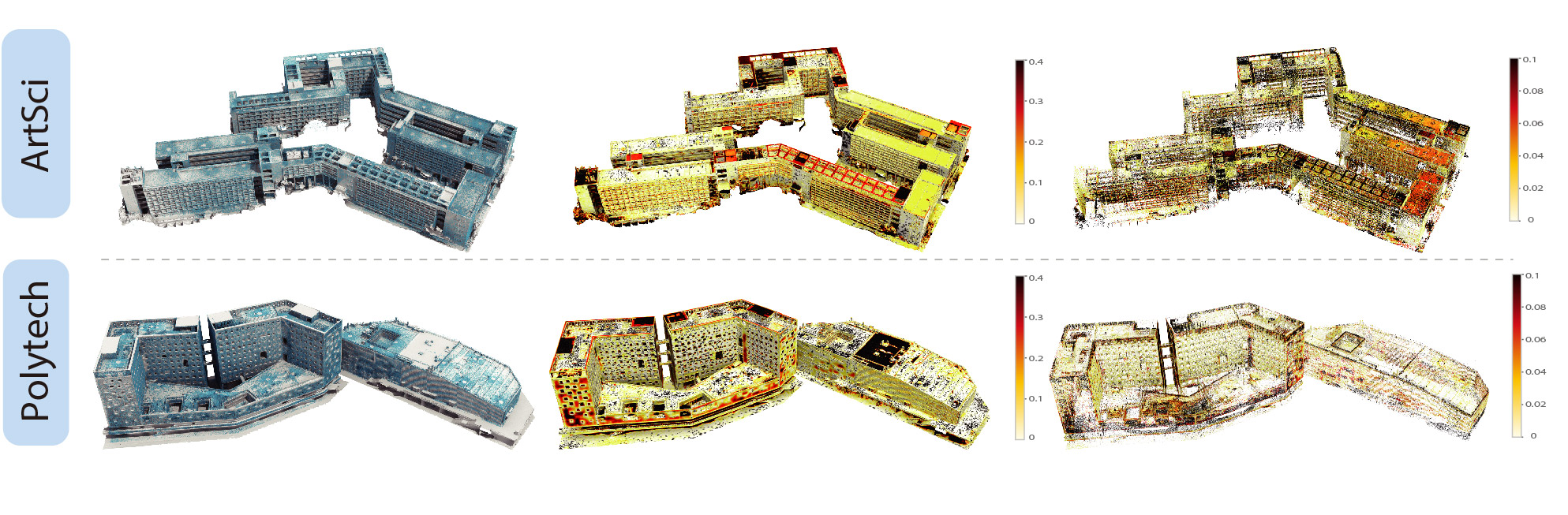}
    \caption{
        The visualization of scanned point clouds and reconstruction error maps for real scene ArtSci and Polytech.
    }
    \label{fig:vis_realscan}
\end{figure}

The synthetic scenes come with ground-truth meshes for evaluation.
For the real scenes, we scan the entire building with high-precision LiDAR scanners loaded with GPS localization devices. The point clouds are then registered with each other, resulting in a high-precision LiDAR scan of the whole building.

The LiDAR scanner is Trimble X7 with self-calibration and self-registration techniques. 
The ranging noise is $0.5mm$, the ranging accuracy is $2mm
$, the angular accuracy is $21''$, and the accuracy of 3D points is $1.5mm$ at $10m$ and $2.4mm$ at $20m$. Each scan, including self-calibration, takes 2 minutes and 34 seconds. 
For real scene Polytech, the overall error of registration is $6mm$; for real scene ArtSci, the overall error of registration is $3mm$.

Fig.~\ref{fig:vis_realscan} shows the scanned point clouds, the accuracy maps, and the completeness maps for both the real scene ArtSci and Polytech.

\subsection{UAV Capturing Cost}
\label{sec:benchmark:cost}

\begin{figure}[!t]
    \centering
    \includegraphics[width=\linewidth]{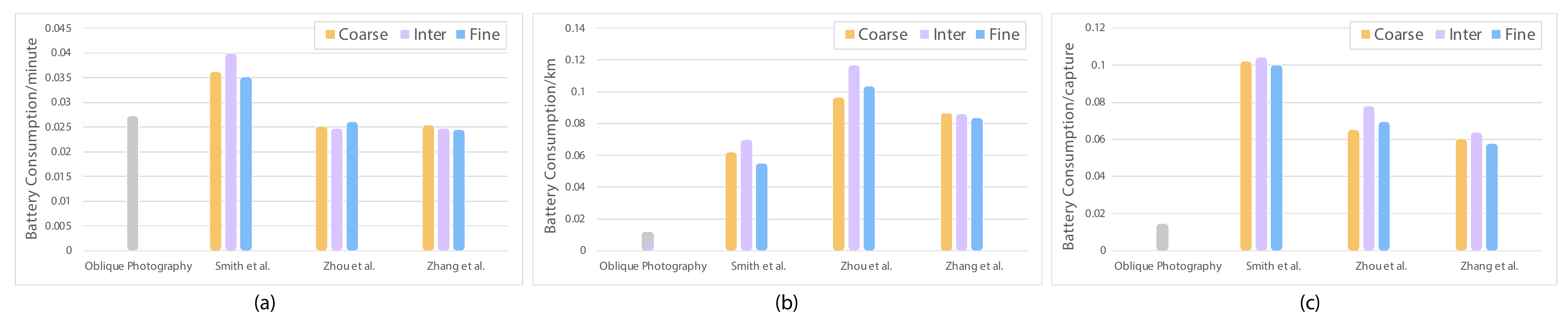}
    \caption{
        Battery consumption of different methods with different proxies on the real scene ArtSci.
        (a): battery consumption per minute (\%/minute) $\downarrow$;
        (b): battery consumption per $Km$ ($10^{-3}\%/km$) $\downarrow$;
        (c): battery consumption between two captures ($10^{-2}$\%) $\downarrow$.
    }
    \label{fig:cost}
\end{figure}

Along with the length of the flight path, the efficiency of the path, such as the total turning angles, affects the overall energy consumption. The drone consumes more energy when it accelerates and decelerates near the turns.

Fig.~\ref{fig:cost} shows the battery consumption statistics of the flight paths planned with different methods on the real scene ArtSci.
As we can see in this figure, the capturing cost, or the efficiency of the flight path, is mainly affected by the path pattern.
Since oblique photography has the simplest path (Fig.~\ref{fig:paths}), it has nearly the lowest battery cost. Generally, the method proposed by Zhang et al.~\cite{zhang2021continuous} has a lower battery consumption than the other two methods, since they explicitly optimize the path efficiency in their cost function.

\subsection{Aerotriangulation Error}
\label{sec:benchmark:reg}

\begin{figure}[!t]
    \centering
    \includegraphics[width=\linewidth]{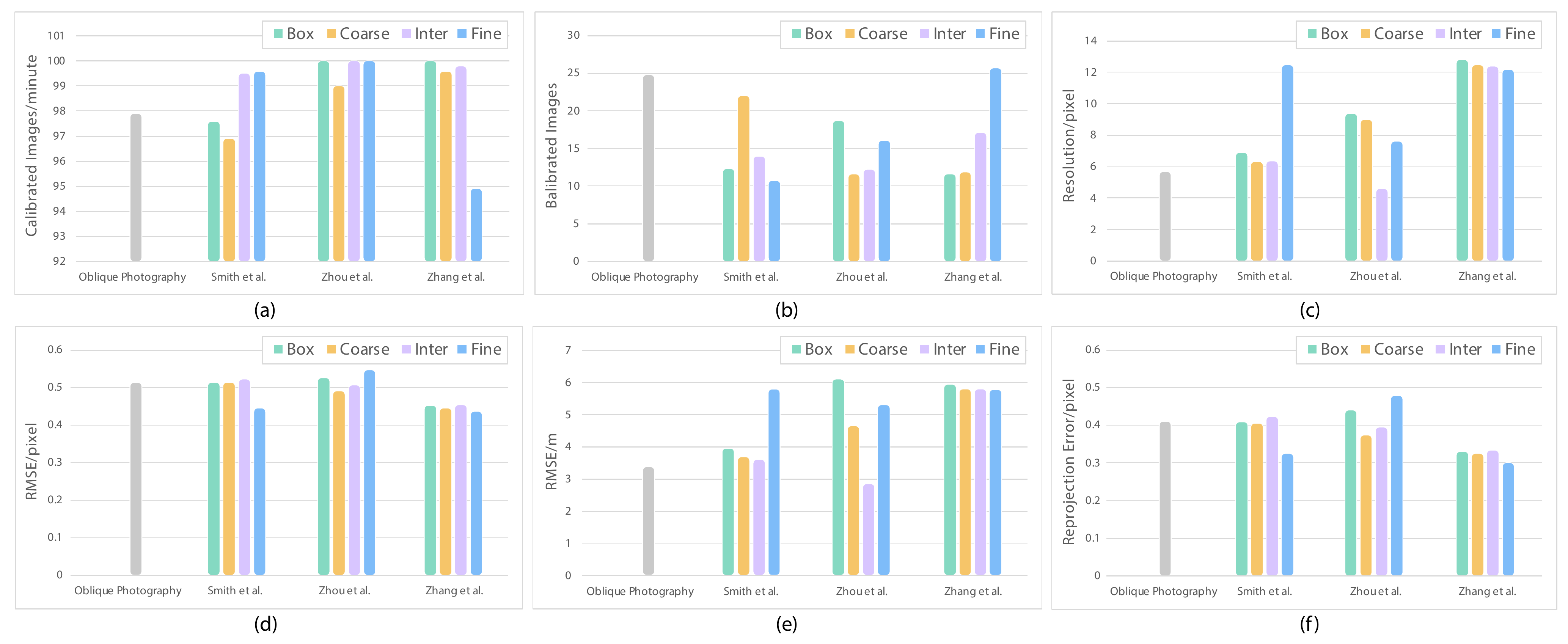}
    \caption{
        Aerotriangulation error of different methods with different proxies on the synthetic scenes Town.
        (a): calibrated images per minute (\#/m) $\uparrow$;
        (b): the rate of succssefully calibrated images (\%) $\uparrow$;
        (c): average resolution per pixel ($1e^{-3}m/pixel$) $\downarrow$;
        (d): root mean square error in pixel ($1e^{-3}$pixel) $\downarrow$;
        (e): root mean square error in meter ($1e^{-3}m$) $\downarrow$;
        (f): reprojection error (pixel) $\downarrow$.
    }
    \label{fig:tri_proxy}
\end{figure}

Before reconstructing the scenes, a process named aerotriangulation, which is a triangulation with aerial images is first performed on the captured images to determine 
the pose of the cameras and to obtain a sparse point cloud of the environment.

Fig.~\ref{fig:tri_proxy} shows the statistics of the aerotriangulation error on different proxies with the overlap as 90\% tested on the scene Town.
As indicated in Fig.~\ref{fig:tri_proxy}, the aerotriangulation results of the method proposed by Smith et al.~\cite{smith2018aerial} and the method proposed by Zhou et al.~\cite{zhou2020offsite} are quite sensitive to the different levels of proxies. For the method proposed by Zhang et al.~\cite{zhang2021continuous}, the RMS-pixel (root mean square error in pixel), RMS-meter (root mean square error in meter), and the reprojection error decrease as the proxy go finer. 
However, compared to Zhou et al.~\cite{zhou2020offsite}, some images captured with this method are not calibrated well. The aerotriangulation results of Zhou et al.~\cite{zhou2020offsite} and Zhang et al.~\cite{zhang2021continuous} have lower RMSE in meter than Smith et al.~\cite{smith2018aerial}. The images captured by oblique photography are also well-calibrated and have a low aerotriangulation error as the images are overlapped with each other strictly.

\subsection{Reconstruction Accuracy and Completeness}
\label{sec:benchmark:recon}

\begin{figure}[!t]
    \centering
    \includegraphics[width=\linewidth]{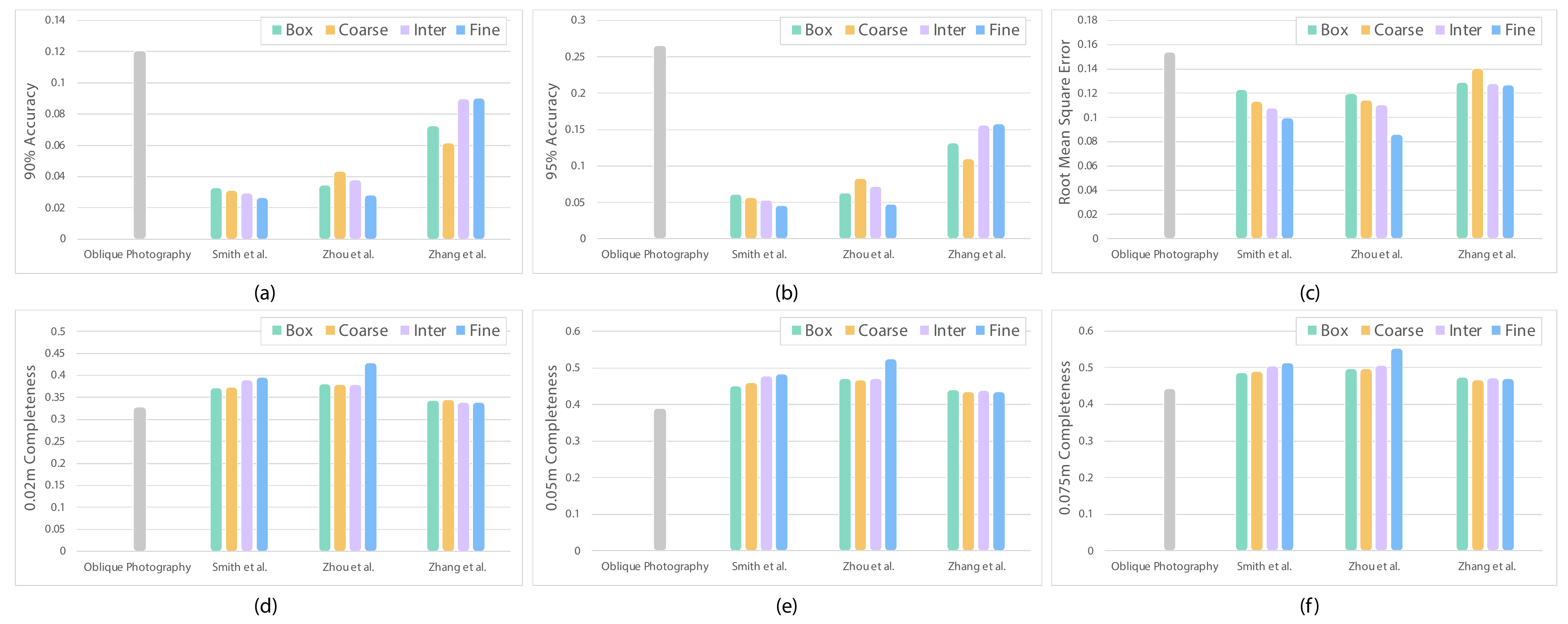}
    \caption{
        Reconstruction error of different methods with different proxies on the synthetic scene School.
        (a): 90\% accuracy ($m$) $\downarrow$;
        (b): 95\% Accuracy ($m$) $\downarrow$;
        (c): root mean square error  ($m$) $\downarrow$;
        (d): 0.02$m$ completeness (\%) $\uparrow$;
        (e): 0.05$m$ completeness (\%) $\uparrow$;
        (f): 0.075$m$ completeness (\%) $\uparrow$.
    }
    \label{fig:recon_proxy}
\end{figure}

We evaluate the reconstruction results of the four planners with different proxies and the results of reconstruction accuracy and completeness are shown in Fig.~\ref{fig:recon_proxy}. The evaluation is performed on scene School with 90\% overlap.

The value of 90\% or 95\% accuracy $x$ means that for all the closest points of the vertices of the reconstruction model on the ground-truth model, 90\% or 95\% of them have a distance less than $x$.
The value of $0.02m$, $0.075m$, or $0.075m$ completeness $x\%$ means that for all the closest points of the vertices of the ground-truth model on the reconstruction model, $x\%$ of them have a distance less than $0.02m$, $0.05m$, or $0.075m$. A smaller 90\% and 95\% accuracy value mean higher accuracy and a larger $0.02m$, $0.05m$, or $0.075m$ completeness value means higher completeness.

As shown in Fig.~\ref{fig:recon_proxy}, both the accuracy and the completeness of reconstructed results of the methods proposed by Smith et al.~\cite{smith2018aerial} and Zhou et al.~\cite{zhou2020offsite} increase as the proxy goes finer. The accuracy and completeness of the results by Zhang et al.~\cite{zhang2021continuous} are not quite consistent with the proxy.

\begin{figure}[!t]
    \centering
    \includegraphics[width=\linewidth]{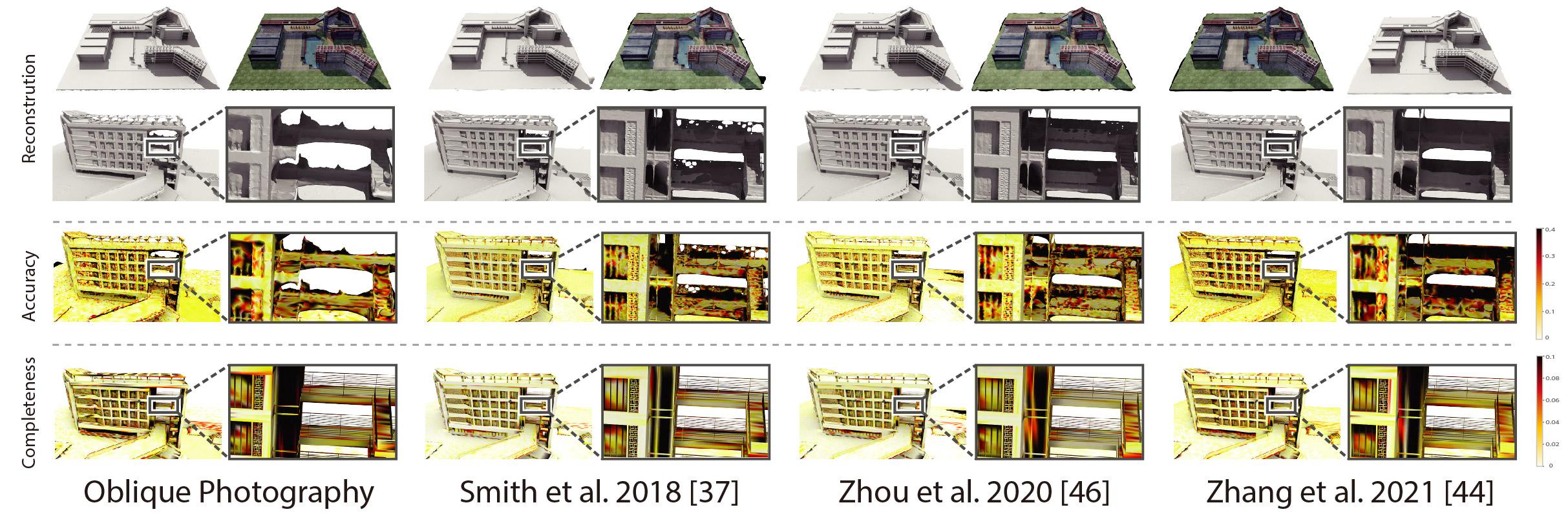}
    \caption{
        Visual comparisons of the resulting 3D reconstruction and the corresponding reconstruction error produced by different methods.
        A higher value means lower accuracy and less completeness for the second and the third rows.
    }
    \label{fig:vis_recon_error}
\end{figure}

The visualization of the reconstructed results and reconstruction error is shown in Fig.~\ref{fig:vis_recon_error}.
The values for accuracy and completeness are clamped to $0 \sim 0.04$ and $0 \sim 0.1$.
As one could expect, the complex geometry and high occlusion induce lower accuracy and completeness of reconstruction.

In general, the method proposed by Smith et al.~\cite{smith2018aerial} and Zhou et al.~\cite{zhou2020offsite} get higher accuracy and completeness compared to oblique photography and Zhang et al.~\cite{zhang2021continuous}. However, the paths produced by the method proposed by Zhang et al.~\cite{zhang2021continuous} have higher quality and consume less energy.

\subsection{Comparison of Different Planners}
\label{sec:benchmark:conclusion}

The path generated with oblique photography simply follows a Zig-Zag pattern thus the target scene is completely covered and the captured images are well calculated. As the baseline planner, oblique photography has the lowest energy cost among all the four planners but results in a roughest reconstruction. The reconstruction error mainly comes from the occlusion between different buildings and other objects scince it can not dive into the space between them.
Both the mothod propose by Smith et al.~\cite{smith2018aerial} and Zhou et al.~\cite{zhou2020offsite} get a much higher quality reconstruction than oblique photography. In general, the mothod proposed by Zhou et al.~\cite{zhou2020offsite} cost less energy compared to the Smith et al.~\cite{smith2018aerial}. For the mothod proposed by Zhang et al.~\cite{zhang2021continuous}, although they get a higher reconstruction error than Smith et al.~\cite{smith2018aerial} and Zhou et al.~\cite{zhou2020offsite}, the battery consumption is reduced due to the continuity of the generated paths.

\section{Simulator and Applications}
\label{sec:app}

Although there are 3D instance segmentation datasets, e.g., S3DIS~\cite{s3dis}, ScanNet~\cite{scannet} and SceneNN~\cite{scenenn}, they are all collected from indoor scenes and still not enough for deep learning based methods. Please note that there is basically no decent dataset for learning 3D building instance segmentation in spacious outdoor scenes, especially for complicated urban regions.

\begin{figure}[!t]
    \centering
    \includegraphics[width=\linewidth]{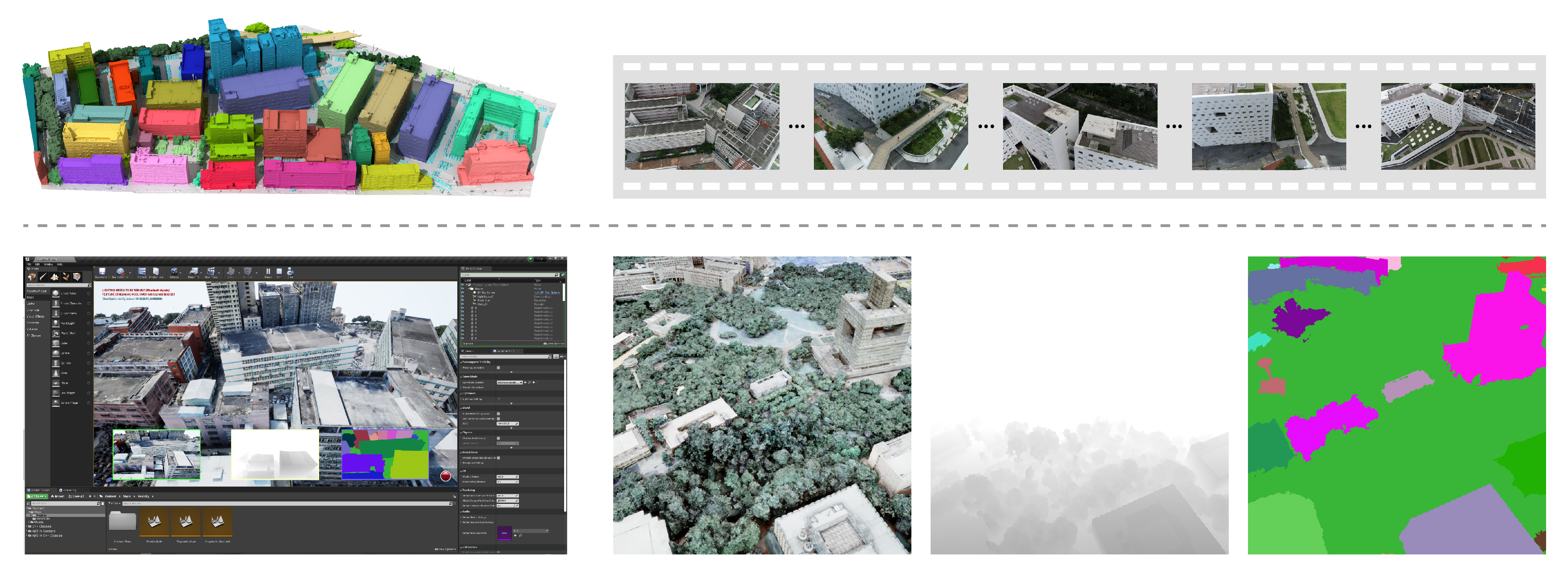}
    \caption{UrbanScene3D also provides the building instance ID for each environment (top left) , 4K aerial videos that are aimed at the real scene acquisition (top right), and a simulator built on Unreal Engine and AirSim (bottom).}
    \label{fig:simulator}
\end{figure}

In this context, our released UrbanScene3D provides rich, large-scale urban scene building annotation data for 3D instance segmentation research. To segment and label 3D architectures, we manually extract all single building models from the entire scene model. 
Every building is then assigned a unique label, forming an instance segmentation map; see the top right of Fig.~\ref{fig:simulator} for an example. 
The 3D textured models with instance segmentation labels in UrbanScene3D allow users to obtain all kinds of data they would like to have: instance segmentation map, depth map in arbitrary resolution, 3D point cloud/mesh in both visible and invisible places, etc. 
UrbanScene3D also offers 4K captured aerial videos in some specific real scenes aimed for 3D reconstruction; see the top left of Fig.~\ref{fig:simulator}. 
Together with high-precision laser scans as ground-truth, these data can be effectively used to train and evaluate various SLAM algorithms.

Moreover, with UrbanScene3D, users can also simulate the robots (cars or drones) to test a variety of autonomous tasks in the proposed city environments. The gravity, inertia and collision can be handled by the physical engine of Airsim. Thus, users can easily generate highly realistic data for many tasks, such as depth estimation, autonomous navigation, and novel view synthesis. Meanwhile, both the lighting condition and the weather of each urban scene can be manipulated by users too. That is, users are able to simulate a rainy night campus or a foggy morning campus; see e.g., the bottom of Fig.~\ref{fig:simulator}. Such data endowed with large diversity would reduce the discrepancy between the simulated and real-world environments, and hence increase the generalization of proposed algorithms.

\section{Conclusion and Future Work}

We present a large-scale dataset, \emph{UrbanScene3D}, which offers rich data annotations and a wide variety of observations of six representative environments. The corresponding reconstruction results and the ground-truth models/scans can be used to evaluate path planning and MVS algorithms. Besides, the proposed simulator allows users to further explore and capture urban scenes in various data patterns with different lighting/weather conditions. The release of UrbanScene3D would largely benefit the community. 

In the future, we plan to do high-level geometric descriptions in the dataset, such as 3D structural points, cross-sectional profiles, wire-frames, or plane segments, etc., to support further research in both computer vision and computer graphics. UrbanScene3D will constantly grow to make more contributions to the data-driven study.

\paragraph*{\textbf{Acknowledgements.}}
This work was supported in parts by NSFC (62161146005, U21B2023, U2001206), GD Talent Program (2019JC05X328), DEGP Key Project (2018KZDXM058, 2020SFKC059), Shenzhen Science and Technology Program (RCJC20200714114435012, JCYJ20210324120213036), and Guangdong Laboratory of Artificial Intelligence and Digital Economy (SZ).

\clearpage
\bibliographystyle{splncs04}
\bibliography{UrbanScene3D}
\end{document}